\documentclass[11pt]{article}
\usepackage{listings}
\usepackage{float}
\usepackage{xurl}
\usepackage{hyperref}
\usepackage{fancyvrb}
\usepackage{booktabs}
\usepackage{amssymb}
\usepackage[preprint]{acl}
\usepackage{enumitem}
\usepackage{amsmath}
\usepackage{makecell, multirow, array} 
\usepackage{times}
\usepackage{latexsym}
\usepackage{tabularx}
\usepackage[T1]{fontenc}

\usepackage[utf8]{inputenc}

\usepackage{microtype}

\usepackage{inconsolata}

\usepackage{graphicx}

\title{Soft Contextualized Encoder For User Defined Text Classification}

\author{Charu Maheshwari \\
  University of Cambridge \\
  \texttt{cm2115@cam.ac.uk} \\\And
  Vyas Raina \\
  APTA \\
  \texttt{vyas@apta.chat} \\}

\begin{document}
\maketitle
\begin{abstract}
User-Defined Text Classification (UDTC) considers the challenge of classifying input text to user-specified, previously unseen classes, a setting that arises frequently in real-world applications such as enterprise analytics, content moderation, and domain-specific information retrieval. We propose a soft-contextualized encoder architecture for UDTC which contextualizes each candidate label with the label set and a static soft prompt representation of the input query. Training on diverse, multi-source datasets enables the model to generalize effectively to zero-shot classification over entirely unseen topic sets drawn from arbitrary domains. We evaluate the proposed architecture both on held-out in-distribution test data and on multiple unseen UDTC benchmarks. Across datasets, the model achieves state-of-the-art performance, consistently outperforming or matching the baselines.
\end{abstract}

\section{Introduction}
Text classification is a standard task in natural language processing: given an input text, the system must assign it to one of several categories. Typically, text classification relies on Transformer architectures where labels are treated as fixed numerical indices within a task-specific head ~\citep{fields2024survey, verma2023comparison, kora2023transformersreview}.

However, many real-world applications require a more flexible setting in which the label set is specified at inference time, and both the number and semantics of the labels may vary across queries. This problem, referred to as \textbf{User-Defined Text Classification (UDTC)}~\citep{chang2008dataless}, poses unique challenges. First, models must interpret arbitrary labels within the context of the entire set, as the most appropriate category for a text often depends on the available alternatives. Second, it must process the two types of inputs - input text and one-word labels that have different semantic structure and length meaningfully. Third, it must generalize to label spaces that were not seen during training. Thus, this warrants for efficient, robust and low latency solutions to UDTC tasks.

A common approach to UDTC is embedding-based similarity, where input texts and label descriptions are embedded independently and compared using measures such as cosine similarity~\citep{sappadla2016semantic}. Despite their simplicity and scalability, such methods often perform poorly in practice ~\citep{8c8b9f52548318c240f4813ca79d9480b29a2dfb}. Static (frozen, non contextualized) embeddings can struggle to capture the fine-grained semantic relations required to resolve label ambiguity and to accommodate the structural and length differences between input texts and labels. Several variants attempt to mitigate these issues by using separate encoders, shared projection spaces, or external semantic resources~\citep{chu2020unsupervised, pappas2018gile, zhang2019semantic}. However, these approaches still produce static label representations, treating labels as fixed targets rather than contextual representations conditioned on the input text or the label set, which can limit their effectiveness in UDTC settings.

Large language model based generative approaches provide an alternative approach for UDTC. Prompt-based classification enables a model to evaluate an input text against candidate labels using natural-language prompts, without requiring parameter updates~\citep{puri2019zeroshot, 97b58563eecaae3df80936068b8793f4a982b373}. While this naturally incorporates label set and broader textual context, its performance is often unstable across domains and highly sensitive to prompt design. Fine-tuning large language models for general classification tasks can improve robustness but remains computationally expensive. Additionally, autoregressive language models exhibit positional biases, making predictions sensitive to the order in which labels are presented and thus violating label-set order invariance ~\citep{fd81018bc72b030545a2d3f3010f3758ec4d48c3, 7c3c9f90e3acc5a0e780b121456a45df8ebed1a0}.

To address these limitations, we introduce a new architecture for UDTC: soft contextualized encoder (Figure~\ref{Encoder}). Labels are represented as order-invariant semantic tokens, eliminating positional bias and preserving label-set invariance. These label tokens are jointly contextualized on the label space through self attention and conditioned on the input text via a soft prompt derived from an external embedding. This design allows for label-set and context-aware classification while minimizing training and inference compute, enabling efficient generalization across unseen label spaces. We focus on architectures at approximately the one-billion-parameter scale and accordingly, evaluate our proposed approach against embedding-based similarity and LLM based generative approaches with comparable parameter counts.

\section{Soft Contextualized Encoder}
\subsection{Task}
User-Defined Text Classification (UDTC) is when a model is able to classify input text into any set of labels at inference time, even those unseen during training, without requiring retraining. Formally, let $x$ be an input text and $K$ be the set of labels:
$\mathcal{T}=\{t_1, t_2, \dots, t_K\}$. $K$ can vary dynamically with each sample (input text $x$). The goal is to learn a function $\mathcal{F}$ that predicts the correct label $\hat{t}$ from the provided set,
\begin{equation}
\hat{t} = \mathcal{F}(x, \mathcal{T}),
\end{equation}
where $\hat{t} \in \mathcal{T}$. 
This classification must be unbiased and invariant to the order in which the labels in $\mathcal{T}$ are presented.

\subsection{Method}
\begin{figure}[h]
    \centering
    \includegraphics[width=0.9\linewidth]{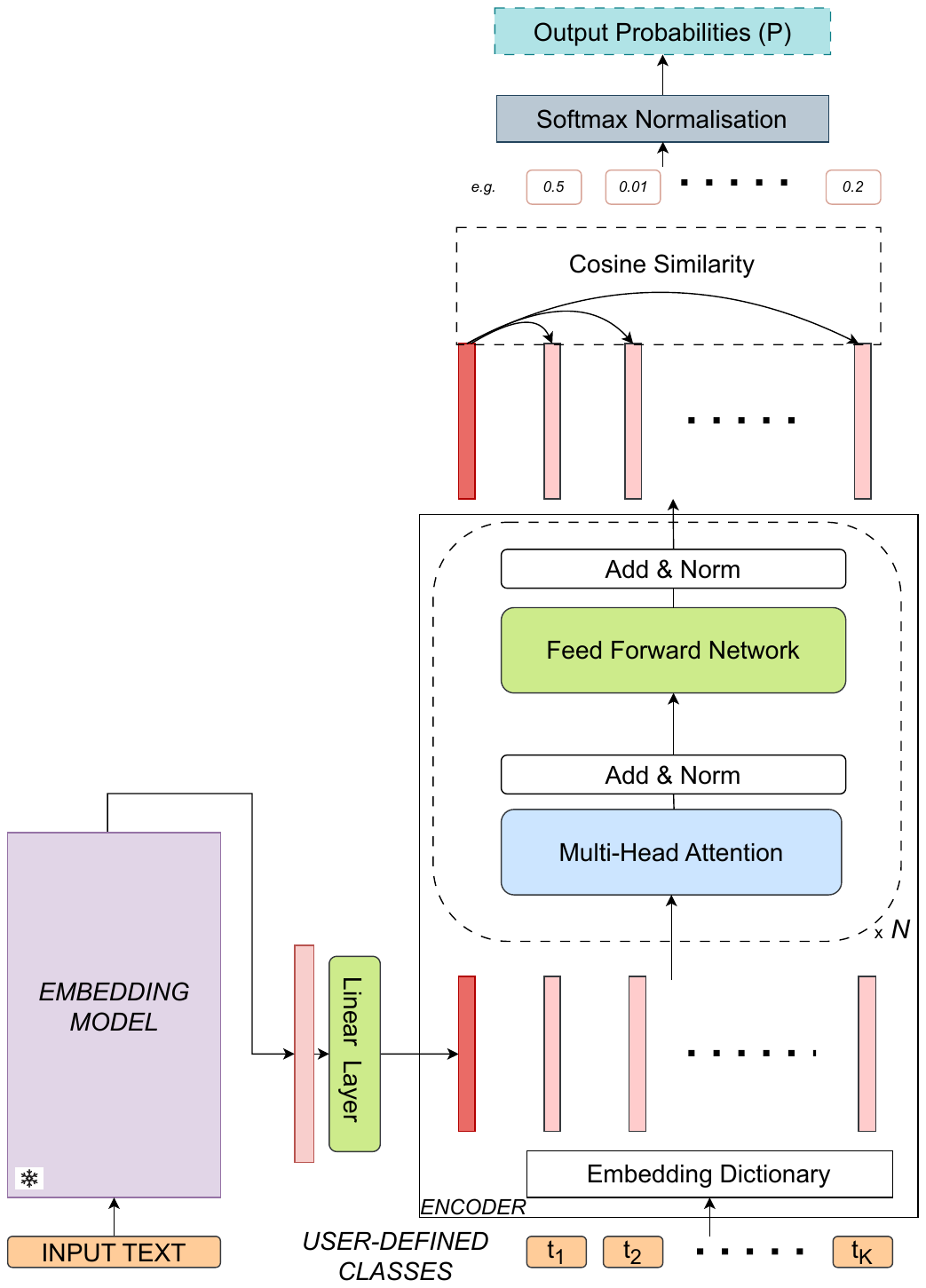}
    \caption{Soft Contextualized Encoder Architecture}
    \label{Encoder}
\end{figure}

\paragraph{Architecture.}
The proposed architecture (Figure ~\ref{Encoder}) comprises of two models, an external, frozen (non-trainable) pretrained embedding model and a trainable encoder. The input text embedding (Query) $\mathbf{q}$ is fed into the text embedding model $g$,
\begin{equation*}
\mathbf{q} = g(x) \in \mathbb{R}^{d_q}.
\end{equation*}
A learned affine map aligns $\mathbf{q}$ to the encoder width:
\begin{equation*}
 \mathbf{q}' \;=\; W\mathbf{q} + \mathbf{b} \;\in\; \mathbb{R}^{d},   
\end{equation*}
with trainable $W\in\mathbb{R}^{d\times d_q}$ and $\mathbf{b}\in\mathbb{R}^{d}$.\\
Each \textbf{label $t_k$} is tokenized to a sequence $\tau_k=(\tau_{k,1},\dots,\tau_{k,m_k})$. We compute token embeddings using the frozen token embedding dictionary $E$ and form an initial label vector, e.g., by mean pooling:
\begin{equation}
\mathbf{h}_k \;=\; \frac{1}{m_k}\sum_{r=1}^{m_k} E(\tau_{k,r})
\quad \in \mathbb{R}^{d}
\qquad (k=1,\dots,K).
\label{Equation:define h}
\end{equation}
In this work, however, we opt for single-token labels.
Let $f(\cdot;\theta)$ be a Transformer encoder ~\citep{vaswani2017attention} without positional encodings i.e., permutation-invariant encoder. We prepend the query vector as a soft prompt and process the set of label embeddings jointly,
\begin{equation*}
\bigl(\mathbf{e}_0,\mathbf{e}_1,\dots,\mathbf{e}_K\bigr)
\;=\;
f\!\left(\,\mathbf{q}',\,\mathbf{h}_1,\,\dots,\,\mathbf{h}_K;\,\theta\right),
\end{equation*}
where $\mathbf{e}_0\in\mathbb{R}^{d}$ is the query’s contextualized output and $\mathbf{e}_k\in\mathbb{R}^{d}$ is the contextualized representation of topic $t_k$. We then score labels by query–label similarity (dot product) and apply the softmax function to simulate a probability distribution, $\hat{\mathbf{p}}$ over the label set,
\begin{equation}
\label{eq:score-softmax}
s_k \;=\; \mathbf{e}_0^\top \mathbf{e}_k,
\qquad
\hat{p}_k \;=\; \frac{\exp(s_k)}{\sum_{j=1}^{K}\exp(s_j)},
\end{equation}
where $k=1,\dots,K$.

\paragraph{Training Objective.} Standard cross entropy loss is used to train the model to minimize the distance between the $\hat{\mathbf{p}}$ from Equation~\ref{eq:score-softmax} and $\mathbf{p}$ (the ground truth, typically one-hot).

We optimize the encoder parameters $\theta$ and the query adaptor $(W,\mathbf{b})$ via stochastic gradient descent over training triples $(x,\mathcal{T},\mathbf{p})$ where the candidate set $\mathcal{T}$ can vary per example. To prevent overfitting, during training, we uniformly sample from a preset of the ground-truth label's $\hat{t}$  synonyms and replace it with the sampled synonym-version.
\paragraph{Attributes.}
The architecture enforces permutation (order) invariance. The encoder uses no positional encodings, hence self-attention is permutation-equivariant over the set $\{\mathbf{h}_1,\dots,\mathbf{h}_K\}$, defined in Equation~\ref{Equation:define h}. Permuting the label inputs yields the same multiset of $\{\mathbf{e}_1,\dots,\mathbf{e}_K\}$ up to the same permutation, while $\mathbf{e}_0$ is unaffected. Equation~\ref{eq:score-softmax} depends only on the unordered set of similarities $\{\mathbf{e}_0^\top\mathbf{e}_k\}$. Hence, the resulting distribution $\hat{\mathbf{p}}$ is invariant to the order in which labels are provided.

Each label is represented by a token embedding input to the encoder, so the label-set cardinality can vary arbitrarily at inference corresponding to the initial number of token embeddings fed in. As the encoder is already pre-trained on a wide range of language, it has the capacity to handle unseen labels during inference, allowing for generalization.

By utilizing the input text encoding as a soft-prompt, we explicitly condition the label representations on the input text $x$. This creates dynamically contextualized embeddings where self-attention ensures awareness of the label space $\mathcal{T}$, while soft-prompting ensures specific input text awareness.
The architecture accounts for the structural disparity between long-form input text and concise, often single-word labels. We essentially train the encoder to learn a domain-agnostic mapping between these two embedding spaces.

\paragraph{Other Approaches.}
We compare the proposed architecture against standard prompt-based LLM classification (see Appendix~\ref{app:baseline-decoder}), as well as embedding-based cosine similarity classification, where the predicted label is selected based on maximum cosine similarity between text and label embeddings (see Appendix ~\ref{app:similarity}).

\section{Experiments}
We focus on models at the one-billion-parameter scale. Architectures with comparable parameter counts are assumed to have similar inference latency, enabling a fair comparison across models with matched inference cost.
\subsection{Setup}
\paragraph{Data.} While this study mainly leverages news classification, the architecture is designed for cross-domain versatility. By training on diverse, multi-source datasets, the model can be effectively deployed for zero-shot inference on entirely unseen label sets from any domain. The datasets used for training the soft-contextualized Encoder are BBC News Archive (BBC) ~\citep{greene06icml}, Huffpost News Category Dataset (Huffpost) ~\citep{misra2022newscategorydataset}, Microsoft News Dataset (MIND) ~\citep{wu-etal-2020-mind}, New York Times Annotated Corpus (NYTimes) ~\citep{sandhaus2008nytimes}. The model's performance is evaluated using both held-out test samples from the training distribution and entirely unseen benchmarks, including AG News (AGNews) ~\citep{zhang2015character}, Yahoo! Answers (Yahoo) ~\citep{zhang2015character}, Okite/Arise News (Arise) ~\citep{okite2021nigeriannews}, and the MN-DS Multilabeled News Dataset (MNDS) ~\citep{petukhova2023mnds}. This assesses the architecture's capacity for both seen and unseen (zero-shot) domain generalization (detailed further in Appendix ~\ref{app:datasets}).
\paragraph{Models.} The Soft Contextualized Encoder (SCE) uses RoBERTa (base) ~\citep{liu2019roberta} as the encoder model for labels and {jina-embedding-v3} ~\citep{sturua2024jina} to get input text embeddings (external embeddings). RoBERTa base is a 110 million parameter model and jina-embedding-v3 built on XLM RoBERTa is 570 million parameters. The total size of proposed architecture is then 680 million parameters. The detailed description of the architecture is given in Appendix ~\ref{app:roberta} and Appendix ~\ref{app:jina}. Meta Llama-3.2-1B-Instruct ~\citep{meta_llama32} is used for Prompt based LLM classification. Further details on its architecture is given in Appendix ~\ref{app:meta-llama}. For the cosine similarity embedding approach, we employ F2LLM ~\citep{zhang2025f2llm}, a popular 1.7B parameter model built upon the Qwen-3 architecture (see Appendix~\ref{app:F2LLM}).\\ 
\paragraph{Training Setup.} In SCE, we initialize the encoder with Roberta base, and train it using Adam Optimizer with a learning rate 1e-5-1e-6 and a weight decay of 1e-4. Jina-embedding-v3 is used to produce the external static text-embeddings with its task-specific `classification' LoRa adapter employed.
The prompt used for classification by off-the-shelf Meta Llama-3.2-1B-Instruct is given in Appendix~\ref{app:prompt:llama-single}.

\subsection{Results and Discussion}
\begin{table}[tb]
\centering
\small
\begin{tabular}{@{}p{0.26\linewidth}ccc@{}}
\toprule
\textbf{Dataset} &
\shortstack{\textbf{SCE}} &
\shortstack{\textbf{Meta-Llama}} &
\shortstack{\textbf{F2LLM}\\\textbf{Embed. Sim}} \\
\midrule
\multicolumn{4}{c}{Unseen Datasets} \\
\midrule
AriseNews        & \textbf{84.66} & 62.90 & 56.40 \\
AGNews           & \textbf{63.22} & 32.11 & 43.29 \\
Yahoo           & \textbf{44.30} & 19.84 & 31.52 \\
MNDS     & 37.67 & 13.00 & \textbf{39.33} \\
\midrule
\multicolumn{4}{c}{Seen Datasets} \\
\midrule
BBC          & \textbf{94.60} & 55.70 & 52.10 \\
Huffpost     &\textbf{ 75.32} & 30.44 & 51.28 \\
MIND        & \textbf{74.75} & 27.80 & 48.60 \\
NYTimes           & \textbf{79.50} & 13.70 & 54.23 \\
\bottomrule
\end{tabular}

\caption{Accuracy(\%) of different methods for seen and unseen UDTC tasks: Proposed approach (SCE).}
\label{table:1}
\end{table}

Despite having fewer parameters, less than half the size of F2LLM and approximately two-thirds the size of Meta-Llama-3.2-1B-Instruct, the proposed Soft Contextualised Encoder (SCE) consistently outperforms or matches these baselines (refer Table ~\ref{table:1}). Importantly, its gains extend beyond tasks observed during training: it achieves superior performance on unseen classification benchmarks such as AriseNews, AGNews, and Yahoo, and remains competitive on MN-DS, with a performance gap of under 2\% (compared to F2LLM). On in-distribution datasets encountered during training, the proposed model yields substantially higher accuracy across all benchmarks. 

\subsection{LLM Fine-tuning Ablation}
An alternative approach to the soft contextualized encoder can be to finetune a general-purpose decoder-based LLM. Therefore, we perform parameter-efficient fine-tuning using Low-Rank Adaptation (LoRA) ~\citep{hu2022lora} on Meta-Llama-3.2-1B-Instruct. To ensure a fair comparison, we configure the LoRA adapter (rank 180) to include 126.8M trainable parameters matching the trainable parameter count in the soft contextualized encoder (approximately the size of RoBERTa-base). We employ the same prompt template used in our primary experiments (Appendix~\ref{app:prompt:llama-single}), with the training objective optimized solely on the target category labels. Further implementation details are provided in Appendix~\ref{app:meta-llama-finetune}. The experimental set-up of the datasets remain the same as earlier.

\begin{table}[htb!]
\centering
\small
\begin{tabular}{@{}p{0.26\linewidth}ccc@{}}
\toprule
\textbf{Dataset} &
\shortstack{\textbf{SCE}} &
\shortstack{\textbf{Meta-Llama}} \\
\midrule
\multicolumn{3}{c}{Unseen Datasets} \\
\midrule
AriseNews        & 84.66 & \textbf{87.92}  \\
AGNews           & 63.22 & \textbf{68.71}  \\
Yahoo  & 44.30 & \textbf{48.40}  \\
MNDS      & 37.67 & \textbf{52.50}  \\
\midrule
\multicolumn{3}{c}{Seen Datasets} \\
\midrule
BBC          & 94.60 & \textbf{97.90}  \\
Huffpost     & 75.32 & \textbf{78.74}  \\
MIND         & 74.75 & \textbf{78.15}  \\
NYTimes           & 79.50 & \textbf{87.42}  \\
\bottomrule
\end{tabular}
\caption{Accuracy(\%) of proposed approach (SCE) against LoRa fine-tuned meta-Llama for UDTC tasks.}
\label{Table 2}
\end{table}
To assess the compute cost for training, we compare the empirical and theoretical training compute times. Empirically, using identical hardware (NVIDIA RTX 6000, 50GB VRAM) and training data, training the RoBERTa encoder takes only 0.006x the time required for one epoch of Meta-Llama-3.2-1B-Instruct. Detailed hyperparameters, configurations, and logged epoch times are provided in Appendix~\ref{app:Comparison-empirical}. These results are corroborated by theoretical calculations in Appendix~\ref{app:Comparison cost}, which show that the incremental epoch cost for the soft-contextualized encoder is approximately 0.005x that of the Llama model. The one-time inference cost for the external embeddings is quickly amortized across multiple epochs. And even when included as an upper bound, Meta-Llama fine-tuning remains computationally expensive (see Appendix~\ref{app:Comparison cost}). Overall the soft-contextualized encoder is dramatically more compute efficient than LoRa-fine-tuning a 1B parameter decoder model.

Although the classification results in Table~\ref{Table 2} show that the soft contextualized encoder achieves marginally lower performance (within 4-15\% of fine-tuned Meta-Llama-Instruct), it requires orders of magnitude less compute to train. Thus though Meta-Llama has better accuracy, its computational overhead makes it an inefficient trade-off for low training-resource settings.  

\section{Conclusion}
UDTC requires classifying text into arbitrary, unseen label sets specified at inference time. In this work, we introduced a soft contextualized encoder for UDTC that represents labels as semantic tokens and jointly contextualizes them with the input text via a soft prompt. Empirical results across seen and unseen benchmarks show that the proposed approach matches or outperforms baselines at the one-billion parameter scale. Future work will further explore the extent of its effectiveness at higher compute scales. 

\section{Limitations}
This work focuses on one-billion parameter scale model, and the empirical results are valid to this setting (requiring low inference time compute), practical to many real world applications. The future work can also compare and extend this architecture to higher order compute settings. The classification considered here is one where the labels carry a semantic meaning known to the pretrained encoder. For example, it can not be extended to grading tasks such as assigning a number between 1 to 5 to some text effectively. A further limitation of this work is that it only focuses on single-label classification as opposed to multi-class classification. For example "positive" and "sport", can not be possibly assigned to the same input text at once.

\bibliography{references}

\appendix

\newpage
~\newpage

\section{Risks and Ethics}
This work presents a methodological contribution to text classification using publicly available benchmark datasets and pre-trained language models. We do not introduce new data or collect personal information. The proposed architecture does not generate new content, or make judgments that could directly affect individuals or groups. Potential down-streams ethical concerns associated with large language models are not specific to our proposed architecture and are inherited from the broader ecosystem of the pre-trained models. Overall, we believe this work has no ethical or societal risks and aligns with responsible research practices in the natural language processing. 

\section{Licensing}
The models used for this work are taken from huggingface under their respective licence: RoBERTa-base is released under the MIT License, jina-embeddings-v3 under the Creative Commons Attribution Non Commercial 4.0 (cc-by-nc-4.0), meta-Llama-3.2-1B-Instruct under Llama 3.2 Community License Agreement(llama3.2) and F2LLM under Apache license 2.0(apache-2.0).
The datasets used for this work are taken from huggingface, kaggle or their original distributional channels or repositories. The HuffPost News category is licensed under Creative Commons Attribution 4.0 International (CC BY 4.0), MIND (Microsoft News Dataset) under Microsoft Research License Terms, New York Times Annotated Corpus(LDC)  under LDC User Agreement for NYT Corpus, Yahoo! Answers Topic Classification under Yahoo Webscope License, okite/arise news under Academic Free License v3.0 (AFL-3.0) and MN-DS (Multilabeled News Dataset) is licensed under Creative Commons Attribution 4.0 International (CC BY 4.0).

\section{Related Work and Existing Research}\label{app:related-work}

\paragraph{Traditional Text Classification.} Classical transformer based text classification assumes a fixed label set and involves training a task or label specific classification head ~\citep{kementchedjhieva2023encoderdecoder} over encoder/encoder-decoder outputs.
Comprehensive surveys and benchmark comparisons such as ~\citet{fields2024survey}, ~\citet{verma2023comparison} and ~\citet{kora2023transformersreview} document how encoder-only architectures (e.g., BERT, RoBERTa) dominate this paradigm, with performance largely governed by encoder depth and input-length handling for context. It often means that labels are treated as specific indices in the classification head rather than the model reasoning over label's natural language meaning.  

\paragraph{Label-agnostic Input-text centric optimization.}
Variants that use twin architectures (similar to ours): two-stream transformer architectures ~\citep{duan2022twostream} or hybrid transformer–LDA topic models ~\citep{tang2023transformerlda} often target multi label classification, but still on a fixed label set. They aim to improve the input text representation, still ignoring label semantics. 

\paragraph{Label-Aware classification constrained to Fixed Label Sets.}
Label specific attention mechanism  to encode the documents/text conditioned on a label with a corresponding classification head exists, but they are restrained to fixed seen classes and don't condition the representation on other labels in the context ~\citep{kementchedjhieva2023encoderdecoder}.

\paragraph{Embedding based approaches.} To encode label's natural language meaning, often embedding based similarity approaches are used. These approaches ~\citep{sappadla2016semantic} are zero shot, and thus generalizable and can be used for multi-label classification. Variants that acknowledge the structural variation between the two input modalities-label and input text, include having different embedding encoders for each ~\citep{chu2020unsupervised} to calculate similarity, labels size independent linear classifier head over different embeddings (text/label) projected on a joint space (GILE) ~\citep{pappas2018gile}. There exists variants that recognize that one-word labels have a different structure that might not be explicit in its meaning or can not be encoded effectively. They use feature augmentation techniques for unseen labels based on knowledge graphs, class description or class hierarchies/taxonomy ~\citep{zhang2019semantic} or integrate unseen classes on the graph dynamically to generate embeddings using Graph Neural Networks ~\citep{zhang2019semantic}. A limitation of these approaches is that they do not produce context-aware label embeddings; instead, they treat labels as static targets rather than representations conditioned on the text or other labels in the label set. 

Other embedding-based extensions include document–graph encoders. In ~\citet{tran2021bertgcn}, both BERT and GCN are trained jointly. The label influence is spread over the BERT embeddings nodes in the graph to counter the structural variations.

\paragraph{Generalizable, label semantic aware classification.} Some multi-label approaches such as ~\citet{chu2020unsupervised} and  ~\citet{halder2020tars} encode label-text pair together and train architectures to get a binary classification for the given pair. However, these approaches fail to take into account the context of the whole label set/space.

\paragraph{Language models based generative approaches.} Prompt based approaches that leverages Large Language models to generate appropriate label(s) remain popular ~\citep{puri2019zeroshot, kementchedjhieva2023encoderdecoder}. While it respects the broader label space and input text context, often the performance is not competitive and the fine-tuning LLMs for particular tasks such as general classification remains expensive. Moreover, because language models are inherently position biased, they no longer remain invariant to order in which they see the label set ~\citep{fd81018bc72b030545a2d3f3010f3758ec4d48c3,7c3c9f90e3acc5a0e780b121456a45df8ebed1a0}. 

\paragraph{Label space and text aware classification lacking Cross Domain Generalization.} ~\citet{kementchedjhieva2023encoderdecoder} presents one optimal configuration where they condition label representations on label space and text. The encoder encodes the text and the decoder is fed labels-as semantic tokens at fixed positions. Classification is done using a binary classifier head at each position of output token embedding. However, this has a major drawback that it is not generalizable to unseen classes at inference time. 

\section{Prompt based LLM classification}\label{app:baseline-decoder}
Consider a causal decoder LLM with vocabulary $\mathcal{V}$ and next-token
logits $\mathbf{z} \in \mathbb{R}^{|\mathcal{V}|}$ conditioned on a prompt.  
We construct a prompt of the form:
\begin{quote}
\small
\texttt{Here is a sentence: } \(x\)\texttt{. Which topic does it belong to from the set \{}%
\(t_1\)\texttt{, }\(t_2\)\texttt{, }\dots\texttt{\}? Return only the topic.}
\end{quote}

After a single forward pass, the model outputs
$\mathbf{z} = f(x) \in \mathbb{R}^{|\mathcal{V}|}$.  
Assume each topic label $t_k$ is represented as a \emph{single token}
in the vocabulary, i.e., $t_k \in \mathcal{V}$.  
We extract the logits corresponding directly to these topic tokens,
yielding the subset $\bigl(z_{t_1}, \dots, z_{t_K}\bigr)$, and compute:
\begin{equation}
\label{eq:llm-softmax}
\hat{p}_k \;=\;
\frac{\exp(z_{t_k})}
     {\sum_{j=1}^{K}\exp(z_{t_j})}
\qquad (k = 1,\dots,K).
\end{equation}

The predicted topic is the one with the highest probability $\hat{p}_k$.

\paragraph{Generalisability, but no permutation invariance.}
Prompt based classification is generalisable to unseen labels at inference time and the classification is text and label space aware. However, because of the inherent positional bias through positional embeddings, the classification is not invariant to order in which the list of labels appear in its prompt.

\section{Cosine Similarity Embedding Based Classification}\label{app:similarity}
Let an embedding model map the input text to a vector representation 
$\mathbf{z} \in \mathbb{R}^d$, and each topic label 
$t_k \in \mathcal{T} = \{t_1, t_2, \dots, t_K\}$ 
to an embedding 
$\mathbf{l}_k \in \mathbb{R}^d$.
We compute a compatibility score between the text and each topic as
\[
s_k = \mathbf{z}^\top \mathbf{l}_k .
\]
These scores are normalized into a probability distribution over 
topics using a softmax:
\[
\hat{p}_k = \frac{\exp(s_k)}{\sum_{j=1}^{K} \exp(s_j)} .
\]
The predicted topic is then the one with the highest probability:
\[
\hat{y} = \arg\max_{k \in \{1,\dots,K\}} \hat{p}_k .
\]

\paragraph{Generalizable, but lacking label-context awareness and structural robustness.}
While the embedding based approach is highly flexible to different inputs, it fails to recognize the structural difference between one-word label and long input text-document modes. 
The representation of each label is statically independent and has no label space or text influence.

\section{Datasets}\label{app:datasets}
Note: For the purpose of this study, to make sure that the topics/labels are one token wide for the soft-contextualized Encoder (RoBERTa) tokenizer - the original labels were modified. To maintain statistical significance, the label sets for certain datasets were refined by either merging fine-grained categories into coarser labels or removing classes with insufficient sample sizes. 
\subsection{Training datasets and \textit{Seen Test} datasets } The datasets used for training, validation and (seen test) are listed below.

\paragraph{BBC News Archive.} ~\citep{greene06icml} It consists of 2225 articles published on the BBC News website during 2004-2005. Taken from huggingface "SetFit/bbc-news". Training - 1.23k (split for the study: train - 1k, validation - 223). Test - 1k. Consists of categories ["tech", "business", "sport", "entertainment", "politics"] which remained unchanged for the scope of this study.  
\paragraph{Huffpost.} ~\citep{misra2022newscategorydataset} This version of the dataset consists of around 210k news headlines from 2012 to 2022 from Huffpost. Taken from Kaggle. Divided into 5000 train, 2000 validation and 5000 test samples. The headline + short description is treated as the text. The original categories ["WELLNESS", "POLITICS", "ENTERTAINMENT", "TRAVEL", "STYLE \& BEAUTY", "PARENTING", "FOOD \& DRINK", "BUSINESS", "SPORTS", "WORLD NEWS"] are mapped to ["wellness", "Politics", "Entertainment", "Travel", "Style", "parenting", "Food", "Business", "Sports", "World"]\footnote{To ensure model robustness, we removed the 'World' category after observing a high tendency for overfitting. The label proved too ill-defined; its meaning shifted across different regional news sources within the training distribution, leading the model to rely on spurious geographical correlations rather than distinct thematic features.}

\paragraph{Mind-Microsoft News Dataset.}~\citep{wu-etal-2020-mind} Collected from the user logs of Microsoft News, it contains 160k+ articles. \footnote{Taken from Microsoft Research original website.} 7000 samples were taken from train set (further divided into 5000 train and 2000 validation samples) and 2000 samples were taken from the test set. 
The original categories ["middleeast", "northamerica", "news", "video", "kids", "games"] were dropped. ["movie", "tv"] were combined to ["entertainment"]. The categories used for this study are  ["travel", "health", "Food", "sports", "weather", "finance", "entertainment", "music", "lifestyle", "autos"].

\paragraph{NYTimes.} New York Times Annotated Corpus. ~\citep{sandhaus2008nytimes} It contains over 1.8 million articles published between 1987 and 2007 and is detailed, using New York Times Index Subject Headings. 5000 samples are used for training, 1100 for validation and 2600 for test. The original label ["Opinion"] is removed, ["Real Estate", "Your Money"] are renamed to ["Economy"]. The category-names used are [   "Television", "Movies", "Sports", "Music", "Theater", "Media", "Art","Style", "Food", "Health", "Science", "Travel", "Economy"]

\begin{table}[h]
\centering
\begin{tabularx}{\columnwidth}{l X r}
\hline
\textbf{Type} & \textbf{Composition} & \textbf{Total} \\ \hline

Validation & Huffpost(2000) + BBC News(225) + MIND(2000) + NYTimes(1100) & 5325 \\ \hline

Train & Huffpost(5000) + BBC News(1000) + MIND(5000) + NYTimes(5000) & 16000 \\ \hline
\end{tabularx}
\caption{Dataset Description For Training and Validation}
\end{table}

\subsection{Unseen Test datasets}

\paragraph{Yahoo Answers Topic Classification.} ~\citep{zhang2015character} Taken from Kaggle. 60k test samples - 5000 used for this study. Question title and Question content forms the text. Original labels ["Society \& Culture", "Science \& Mathematics", "Health", "Education \& Reference", "Computers 
\& Internet", "Sports", "Business \& Finance", "Entertainment \& Music", "Family \& Relationships", "Politics \& Government"] were mapped to ["Society","Science","Health","Education" ,"Computers","Sports","Business","Entertainment",
"Relationships","Government" ].
\paragraph{AG\_News.} ~\citep{zhang2015character}The "AG's Corpus of News Articles" collection is taken from huggingface "SetFit/ag\_news" test(7.6k). AG\_News had originally 4 categories ['Sci/Tech', 'World', 'Sports', 'Business'] which are mapped to ['Science', 'World', 'Sports', 'Business'].
\paragraph{Okite/Arise News.} ~\citep{okite2021nigeriannews} It is an African News Classification Dataset. Taken from Huggingface "okite97/news-data". Test - 828 samples. The original categories ["business", "sports", "politics", "health", "tech", "entertainment"] remain unchanged. 
\paragraph{MNDS-Multilabeled News Dataset.}\footnote{Taken from -'https://zenodo.org/record/7394851/files/MN-DS-news-classification.csv?download=1'} ~\citep{petukhova2023mnds}. The MN-DS is a newer dataset consisting of 10917 articles. Unique for its hierarchical nature, it consists of 17 first level categories and 109 second level categories. The headline and abstract forms the text. 600 samples were taken for testing with equal number of each category . The original categories ["lifestyle and leisure", "arts, culture, entertainment and media", "human interest", "economy, business and finance" ] were dropped and the category-names used were ["society", "science", "politics","sport", "religion", "war", "health", "education", "environment", "crime", "disaster", "weather"]. 

\section{Prompts}
\subsection{meta-LLama-3.2-1B-Instruct prompt}\label{app:prompt:llama-single}
\begin{lstlisting}
"""
You are given a text excerpt and a list of categories.
Your task is to select the best category from the list of categories to classify the given text excerpt.

### Note:
Your selected category for the text excerpt must match exactly -case, plural/singular, and spelling-with one of the options in the CATEGORIES list. Use the category string exactly as it appears in CATEGORIES.

## TEXT_EXCERPT:
{text_excerpt}

## CATEGORIES:
{categories}

## SCHEMA:
SelectedCategory

### Note: Your output must strictly be a single token which is the category selected from the list of categories. Do not give verbose reasonings giving explanations around it. Do not give python reasoning. Just a single token output. Now generate the output.
"""
\end{lstlisting}

\section{Generic notation used to denote model architectural variables}
Let $d$ denote the model hidden size, $d_{\text{ff}}$ the intermediate (expanded) dimension in the feed-forward network (FFN),
$n_{\text{f1}}$ the number of layers expanding the dimension from $d$ to $d_{\text{ff}}$,
$n_{\text{f2}}$ the number of layers reducing the dimension from $d_{\text{ff}}$ to $d$,
$H$ the number of attention heads, $d_h$ the head dimension, $L$ denote the number of layers/transformer blocks, $B$ the batch size,
and $m$ the number of tokens per sequence.

\section{RoBERTa}\label{app:roberta}
RoBERTa$_{\text{Base}}$ has $L = 12$ encoder layers, hidden size $d = 768$, feed-forward dimension $d_{\mathrm{ff}} = 3072$, and $H = 12$ heads, giving approximately $110$M parameters. 
RoBERTa has $n_{f1} = 1$ expansion and $n_{f2} = 1$ contraction layer in the FFN. ~\citep{liu2019roberta,huggingface_roberta_docs}.

\section{jina-Embedding-v3}\label{app:jina} 
jina-embeddings-v3
text embedding model ~\citep{sturua2024jina} is 570 million
parameters, and achieves state-of-the-art performance on multilingual data and long-context
retrieval tasks, supporting context lengths
of up to 8192 tokens. The model includes a set of task-specific Low-Rank Adaptation (LoRA) adapters of rank 4 (employed in its attention layers) to generate high quality embeddings for query-document retrieval, clustering, classification, and text
matching. 
It is built on RoBERTa-XLM (large) and thus has heads $H = 16$, hidden size/output dim $d = 1024$, feed forward dimension $d_{\mathrm{ff}}=4096$, encoder layers $L = 24$, contraction layer in FFN $n_{f2}=1$, expansion layer in FFN $n_{f1}=1$ (with GeLU activation) ~\citep{hf_xlm_roberta_large_config}.

\section{F2LLM}\label{app:F2LLM} 
F2LLM ~\citep{zhang2025f2llm} is a direct contrastive fine-tuning of the existing Qwen3 foundation model without architectural modification. F2LLM-1.7B is used for this study. Qwen-3-1.7B is a decoder only model with $H = 16$, hidden size $d = 2048$, feed forward dimension $d_{\mathrm{ff}}=6144$, encoder layers $L = 28$, contraction layer in FFN $n_{f2}=1$, expansion layer in FFN $n_{f1}=2$ (with SwiGLU activation). ~\citep{qwen3technicalreport, qwen3_1p7b_hf}.Empirically, F2LLM achieved higher cosine similarity when using ``document embeddings" plugin for both labels and input text, leading us to adopt this approach over its separate ``query" and ``document embeddings" for calculating cosine based similarities for this work.

\section{Off-Shelf meta-Llama-3.2 1B-Instruct}\label{app:meta-llama}
The feed-forward network (FFN) contains $n_{f1} = 2$ expansion layers and $n_{f2} = 1$ contraction layer. 
The model has heads $H = 32$, hidden dimension $d = 2048$, feed forward dimension $d_{\text{ff}}  = 8192$, number of layers $L = 16$ ~\citep{meta_llama32} ~\footnote{Meta-Llama-3.2-1B-Instruct was chosen for this work because the base 1B model produced overly verbose, and failed to follow the single-token, JSON-formatted classification instructions.}.

\section{Lora Fine tuning Meta-Llama}\label{app:meta-llama-finetune}
The model used is  Meta LLaMA-3.2-1B architecture. The model has a total of $1{,}362{,}626{,}560$ parameters. LoRA adapters with rank $r = 180$  (adding $126{,}812{,}160$ trainable parameters) were applied to the modules  \texttt{q\_proj}, \texttt{k\_proj}, \texttt{v\_proj}, \texttt{o\_proj}, 
\texttt{gate\_proj}, \texttt{up\_proj} and \texttt{down\_proj}. 
These modules are explained below:
\begin{itemize}[noitemsep, topsep=0pt]
    \item The \texttt{gate\_proj}, \texttt{up\_proj} and \texttt{down\_proj} form part of the feed forward network. The input to the feed forward network is passed through two parallel linear layers: the \textbf{gate layer} projection of x, which is then passed through SiLU activation, and the \textbf{up layer} projection of x. Both these results are multiplied element-wise. The result is then passed through another linear projection: \textbf{down layer} to return to the original dimension.

    \item  The modules \texttt{q\_proj}, \texttt{k\_proj}, \texttt{v\_proj}, \texttt{o\_proj} form part of masked grouped-query attention.The query, key and value projection matrices transform the input into the respective query, key and value embeddings for attention to be performed over. \textbf{Orthogonal projection} matrices map the concatenated multi-head attention output back to model hidden dimension.
\end{itemize}

All frozen weights were stored using 4-bit \texttt{nf4} quantization.

\section{Forward pass Flop Analysis}\label{app:forward_pass}
For a general transformer, the cost of a forward pass is generally equal to the cost of matrix multiplications. Let FLOPs be 1 unit of computation \footnote{1 addition + 1 multiplication or simply 1 addition}.

\paragraph{Linear Layer.}
Consider a linear transformation
\[
\mathbf{y} = \mathbf{x}\mathbf{W},
\]
where $\mathbf{x} \in \mathbb{R}^{1 \times d_{\text{in}}}$, 
$\mathbf{W} \in \mathbb{R}^{d_{\text{in}} \times d_{\text{out}}}$, 
and $\mathbf{y} \in \mathbb{R}^{1 \times d_{\text{out}}}$. 
The forward pass involves a single matrix–vector multiplication:

\begin{align*}
\mathbf{y} &= \mathbf{x}\mathbf{W} 
\qquad \text{FLOPs}_{\mathbf{y}} = d_{\text{in}} d_{\text{out}}.
\end{align*}

For $m$ tokens per sequence, multiply by $m$.

\paragraph{Attention Layers.}
Let $\mathbf{Q}, \mathbf{K}, \mathbf{V} \in \mathbb{R}^{m \times d_{h}}$. 
The attention forward computation proceeds as follows:

\begin{align*}
\mathbf{S} &= \mathbf{Q}\mathbf{K}^{\top}
\qquad \text{FLOPs}_{\mathbf{S}} = m^{2} d_{h}, \\[4pt]
\mathbf{A} &= \mathrm{softmax}(\mathbf{S})
\qquad \text{(softmax cost negligible)}, \\[4pt]
\mathbf{Y} &= \mathbf{A}\mathbf{V}
\qquad \text{FLOPs}_{\mathbf{Y}} = m^{2} d_{h}.
\end{align*}

For $H$ attention heads with $d = H d_{h}$, 
the total cost across all heads is:
\[
\text{FLOPs}_{\text{attn}} = 2m^{2}d.
\]

Number of Flops for one transformer block ignoring batch norm costs is 
\begin{table}[H]
\small
\centering
\renewcommand{\arraystretch}{1.25}

\begin{tabularx}{\columnwidth}{
  p{2.6cm}
  p{2.6cm}
  p{1.6cm}
}
\toprule
\textbf{Component} & \textbf{Operation} & \textbf{Forward FLOPs} \\
\midrule

\multirow{3}{2.6cm}{Linear Projections (Attention)}
  & q\_proj: $d \to d$ & $md^{2}$ \\
  & k\_proj: $d \to d$ & $md^{2}$ \\
  & v\_proj: $d \to d$ & $md^{2}$ \\
\midrule

Output Projection
  & o\_proj: $d \to d$ & $md^{2}$ \\
\midrule

\multicolumn{2}{l}{\textbf{Total Attention Linear Projections}}
  & $4md^{2}$ \\
\midrule

\multirow{2}{2.6cm}{Attention Computation}
  & $\mathbf{Q}\mathbf{K}^\top$ & $m^{2}d_h$ \\
  & $\mathbf{A}\mathbf{V}$      & $m^{2}d_h$ \\
\midrule

\multicolumn{2}{l}{\textbf{Sum over Heads}}
  & $2m^{2}d = 2m^{2}H d_h$ \\
\midrule

\multirow{2}{2.6cm}{Feed-Forward (FFN)}
  & Expansion $(d \to d_{\text{ff}})$,\ $n_{\text{f1}}$
  & $m d d_{\text{ff}} n_{\text{f1}}$ \\
  & Contraction $(d_{\text{ff}} \to d)$,\ $n_{\text{f2}}$
  & $m d_{\text{ff}} d n_{\text{f2}}$ \\
\midrule

\multicolumn{2}{l}{\textbf{FFN Total}}
  & $mdd_{\text{ff}}(n_{\text{f1}} + n_{\text{f2}})$ \\
\midrule

\multicolumn{2}{l}{\textbf{Total FLOPs per Transformer Block}}
  &
 
    $4md^{2} + 2m^{2}d+\, m d d_{\text{ff}}(n_{\text{f1}} + n_{\text{f2}})$
  \\
\bottomrule

\end{tabularx}

\caption{Forward pass FLOPs for one Transformer block (standard model).}
\label{tab:flops-block}
\end{table}

Total Number of Forward FLOPs for the model = $L \cdot  \text{FLOP}_{\text{block}}$.
\begin{table}[H]
\small
\centering
\renewcommand{\arraystretch}{1.15}

\begin{tabularx}{\columnwidth}{
  p{2.6cm}
  p{2.6cm}
  >{\raggedleft\arraybackslash}p{1.6cm}
}
\toprule
\textbf{Component} & \textbf{Operation} & \textbf{Forward FLOPs (LoRA)} \\
\midrule

\multirow{4}{2.6cm}{Linear Projections (Attention + LoRA)}
  & q\_proj: $d\!\to\!d$ (rank $r$) & $m(d^{2}+2dr)$ \\
  & k\_proj: $d\!\to\!d$ (rank $r$) & $m(d^{2}+2dr)$ \\
  & v\_proj: $d\!\to\!d$ (rank $r$) & $m(d^{2}+2dr)$ \\
  & o\_proj: $d\!\to\!d$ (rank $r$) & $m(d^{2}+2dr)$ \\
\midrule

\multicolumn{2}{l}{\textbf{Total Attention Linear Projections}}
  & $4m(d^{2}+2dr)$ \\
\midrule

\multirow{2}{2.6cm}{Attention Computation}
  & $\mathbf{Q}\mathbf{K}^{\top}$ & $m^{2}d_{h}$ \\
  & $\mathbf{A}\mathbf{V}$        & $m^{2}d_{h}$ \\
\midrule

\multicolumn{2}{l}{\textbf{Sum over Heads}}
  & $2m^{2}d = 2m^{2}H d_{h}$ \\
\midrule

\multirow{2}{2.6cm}{Feed-Forward (FFN + LoRA)}
  & Expansion $(d\!\to\!d_{\text{ff}})$,\ $n_{\text{f1}}$
  & $m(d d_{\text{ff}} + d r + d_{\text{ff}} r)n_{\text{f1}}$ \\
  & Contraction $(d_{\text{ff}}\!\to\!d)$,\ $n_{\text{f2}}$
  & $m(d d_{\text{ff}} + d r + d_{\text{ff}} r)n_{\text{f2}}$ \\
\midrule

\multicolumn{3}{l}{\textbf{FFN Total}}
$m(d d_{\text{ff}} + d r + d_{\text{ff}} r)(n_{\text{f1}} + n_{\text{f2}})$ \\
\midrule

\multicolumn{3}{l}{\textbf{Total FLOPs per Transformer Block}}\\
\multicolumn{3}{l}{
$4m(d^{2}+2dr) + 2m^{2}d +\, m(d d_{\text{ff}} + d r + d_{\text{ff}} r)(n_{\text{f1}} + n_{\text{f2}})$}\\

\bottomrule

\end{tabularx}

\caption{Forward pass FLOPs for one Transformer block (LoRA fine-tuned).}
\label{tab:flops-lora-block}
\end{table}

\section{Backward pass Flop Analysis}

During backpropagation, the goal is to compute gradients of the loss with respect to:
\begin{itemize}[noitemsep, topsep=0pt,leftmargin=*]
    \item \textbf{Weights} $\mathbf{W}$: $\displaystyle \frac{\partial L}{\partial \mathbf{W}} \text{ denoted as } \partial \mathbf{W}$
    \item \textbf{Inputs (activations)} $\mathbf{x}$: $\displaystyle \frac{\partial L}{\partial \mathbf{x}} \text{ denoted as }  \partial \mathbf{x}$
\end{itemize}

\paragraph{ Linear Layer.}

Consider a linear transformation 
\[
\mathbf{y} = \mathbf{x}\mathbf{W},
\]
where $\mathbf{x} \in \mathbb{R}^{1 \times d_{\text{in}}}$, $\mathbf{W} \in \mathbb{R}^{d_{\text{in}} \times d_{\text{out}}}$, and $\mathbf{y} \in \mathbb{R}^{1 \times d_{\text{out}}}$.

The backward pass involves computing gradients with respect to both $\mathbf{W}$ and $\mathbf{x}$.
\begin{align*}
\frac{\partial L}{\partial \mathbf{W}} &= \mathbf{x}^{T} \frac{\partial L}{\partial \mathbf{y}} \quad  \text{FLOPs}_{\partial \mathbf{W}} = d_{\text{in}} d_{\text{out}} \\
\frac{\partial L}{\partial \mathbf{x}} &= \frac{\partial L}{\partial \mathbf{y}} \mathbf{W}^{T} \quad  \text{FLOPs}_{\partial \mathbf{x}} = d_{\text{in}} d_{\text{out}}
\end{align*}

For $m$ tokens per sequence, multiply by $m$.

\paragraph{Attention Activation Layers.}

Let $\mathbf{Q}, \mathbf{K}, \mathbf{V}\!\in\!\mathbb{R}^{m\times d_h}$.
The Core attention mechanism comprises of the three steps and the cost for each of them is summarised below :- \\
\noindent\textbf{Backward through } $\mathbf{S}=\mathbf{Q}\mathbf{K}^{\top}$:
\[
\begin{aligned}
\frac{\partial L}{\partial \mathbf{Q}} &= \frac{\partial L}{\partial \mathbf{S}}\;\mathbf{K}
&\qquad \mathrm{FLOPs}_{\mathbf{Q}} &= m^{2} d_{h}, \\
\frac{\partial L}{\partial \mathbf{K}} &= \Big(\tfrac{\partial L}{\partial \mathbf{S}}\Big)^{\!\top}\mathbf{Q}
&\qquad \mathrm{FLOPs}_{\mathbf{K}} &= m^{2} d_{h}.
\end{aligned}
\]

\noindent\textbf{Softmax:} $\mathbf{A}=\mathrm{softmax}(\mathbf{S})$ \; (cost lower order; typically neglected).

\noindent\textbf{Backward through } $\mathbf{Y}=\mathbf{A}\mathbf{V}$:
\[
\begin{aligned}
\frac{\partial L}{\partial \mathbf{A}} &= \frac{\partial L}{\partial \mathbf{Y}}\;\mathbf{V}^{\top}
&\qquad \mathrm{FLOPs}_{\mathbf{A}} &= m^{2} d_{h}, \\
\frac{\partial L}{\partial \mathbf{V}} &= \mathbf{A}^{\top}\,\frac{\partial L}{\partial \mathbf{Y}}
&\qquad \mathrm{FLOPs}_{\mathbf{V}} &= m^{2} d_{h}.
\end{aligned}
\]

\noindent\textbf{Per-head total:} $\text{FLOPs}_{\text{attn head}} = 4\,m^{2}d_{h}$ \\[2pt]
\textbf{For all heads:} $\text{FLOPs}_{\text{attn}} = 4\,m^{2}d$, \; where $d = H d_{h}$.

\paragraph{Backward FLOPs: Standard Transformer}
\begin{table}[H]
\small
\centering
\renewcommand{\arraystretch}{1.3}

\begin{tabularx}{\columnwidth}{
  p{2.6cm}
  p{2.6cm}
  >{\raggedleft\arraybackslash}p{1.6cm}
}
\toprule
\textbf{Component} & \textbf{Operation} & \textbf{Backward FLOPs} \\
\midrule

\multirow{4}{2.6cm}{Linear Projections (Attention)}
  & q\_proj: $d \to d$ & $2md^{2}$ \\
  & k\_proj: $d \to d$ & $2md^{2}$ \\
  & v\_proj: $d \to d$ & $2md^{2}$ \\
  & o\_proj: $d \to d$ & $2md^{2}$ \\
\midrule

\multicolumn{2}{l}{\textbf{Total Attention Linear Projections}}
  & $8md^{2}$ \\
\midrule

\multirow{2}{2.6cm}{Attention Computation}
  & $\mathbf{Q}\mathbf{K}^{\top}$ & $2m^{2}d_h$ \\
  & $\mathbf{A}\mathbf{V}$        & $2m^{2}d_h$ \\
\midrule

\multicolumn{2}{l}{\textbf{Sum over Heads}}
  & $4m^{2}d \;=\; 4m^{2}H d_h$ \\
\midrule

\multirow{2}{2.6cm}{Feed-Forward (FFN)}
  & Expansion $(d \to d_{\text{ff}})$,\ $n_{\text{f1}}$
  & $2m d\, d_{\text{ff}}\, n_{\text{f1}}$ \\
  & Contraction $(d_{\text{ff}} \to d)$,\ $n_{\text{f2}}$
  & $2m d_{\text{ff}} d\, n_{\text{f2}}$ \\
\midrule

\multicolumn{2}{l}{\textbf{FFN Total}}
  & $2m d\, d_{\text{ff}}(n_{\text{f1}}+n_{\text{f2}})$ \\
\midrule

\multicolumn{3}{l}{\textbf{Total FLOPs per Transformer Block}}\\
  
\multicolumn{3}{l}{ $8md^{2} + 4m^{2}d
      +\,2m d\, d_{\text{ff}}(n_{\text{f1}}+n_{\text{f2}})
    $} \\
\bottomrule

\end{tabularx}

\caption{Backward pass FLOPs for one Transformer block/layer (standard model).}
\label{tab:backward-flops-block}
\end{table}

\[
\text{FLOPs}_{\text{model}} = L\,\text{FLOPs}_{\text{block}}\]

\paragraph{Backward FLOPs: LoRA Fine-Tuning.}

For LoRA fine-tuning, each linear layer weight is modified as
\[
\mathbf{W} = \mathbf{W} + \mathbf{A}\mathbf{B}, \quad 
\mathbf{A}\!\in\!\mathbb{R}^{d_{\text{in}}\times r},\;
\mathbf{B}\!\in\!\mathbb{R}^{r\times d_{\text{out}}}.
\]
Hence,
\[
\mathbf{Y} = \mathbf{X}(\mathbf{W}+\mathbf{A}\mathbf{B}) = \mathbf{XW}+(\mathbf{X}\mathbf{A})\mathbf{B} = \mathbf{XW}+\mathbf{Z}\mathbf{B}\]
\[
\quad \mathbf{Z}=\mathbf{X}\mathbf{A}.
\]

\begin{itemize}[noitemsep, topsep=0pt,leftmargin=*]
    \item \textbf{Gradients w.r.t. activations:}
    \[
    \partial \mathbf{X}_{W} = \partial \mathbf{Y}\mathbf{W}^{T},
    \partial \mathbf{Z} = \partial \mathbf{Y}\mathbf{B}^{T},
    \partial \mathbf{X}_{A} = \partial \mathbf{Z}\mathbf{A}^{T}.
    \]
    \item \textbf{Gradients w.r.t. LoRA parameters:}
    \[
    \partial \mathbf{A} = \mathbf{X}^{T}\partial \mathbf{Z}, \quad
    \partial \mathbf{B} = (\mathbf{X}\mathbf{A})^{T}\partial \mathbf{Y}.
    \]
\end{itemize}

Total backward FLOPs for one LoRA linear layer:
\[
\text{FLOPs}_{\text{LoRA linear}} = d_{\text{in}}d_{\text{out}} + 2d_{\text{in}}r + 2r d_{\text{out}}.
\]

\begin{table}[H]
\small
\centering
\renewcommand{\arraystretch}{1.3}

\begin{tabularx}{\columnwidth}{
  p{2.6cm}
  p{2.6cm}
  >{\raggedleft\arraybackslash}p{1.6cm}
}
\toprule
\textbf{Component} & \textbf{Operation} & \textbf{Backward FLOPs (LoRA)} \\
\midrule

\multirow{4}{2.6cm}{\makecell[l]{Linear Projections\\(Attention + LoRA)}}
  & q\_proj: $d\!\to\!d$ (rank $r$) & $m(d^{2}+4dr)$ \\
  & k\_proj: $d\!\to\!d$ (rank $r$) & $m(d^{2}+4dr)$ \\
  & v\_proj: $d\!\to\!d$ (rank $r$) & $m(d^{2}+4dr)$ \\
  & o\_proj: $d\!\to\!d$ (rank $r$) & $m(d^{2}+4dr)$ \\
\midrule

\multicolumn{2}{l}{\textbf{Total Attention Linear Projections}}
  & $4m(d^{2}+4dr)$ \\
\midrule

\multirow{2}{2.6cm}{\makecell[l]{Attention\\Computation}}
  & $\mathbf{Q}\mathbf{K}^{\top}$ & $2m^{2}d_h$ \\
  & $\mathbf{A}\mathbf{V}$        & $2m^{2}d_h$ \\
\midrule

\multicolumn{2}{l}{\textbf{Sum over Heads}}
  & $4m^{2}d = 4m^{2}H d_h$ \\
\midrule

\multirow{2}{2.6cm}{\makecell[l]{Feed-Forward\\(FFN + LoRA)}}
  & Expansion $(d\!\to\!d_{\text{ff}})$,\ $n_{\text{f1}}$
    & $m(d d_{\text{ff}} + 2dr + 2d_{\text{ff}}r)n_{\text{f1}}$ \\
  & Contraction $(d_{\text{ff}}\!\to\!d)$,\ $n_{\text{f2}}$
    & $m(d d_{\text{ff}} + 2dr + 2d_{\text{ff}}r)n_{\text{f2}}$ \\
\midrule

\multicolumn{2}{l}{\textbf{FFN Total}}
  $m(d d_{\text{ff}} + 2dr + 2d_{\text{ff}}r)(n_{\text{f1}}+n_{\text{f2}})$ \\
\midrule
\multicolumn{3}{l}{\textbf{FFN Total}}\\
\multicolumn{3}{l}{  $4m(d^{2}+4dr) + 4m^{2}d+\, m(d d_{\text{ff}} + 2dr + 2d_{\text{ff}}r)(n_{\text{f1}}+n_{\text{f2}})$} \\
\bottomrule

\end{tabularx}

\caption{Backward pass FLOPs for one Transformer block (LoRA fine-tuned).}
\end{table}
\[
\text{FLOPs}_{\text{LoRA model}} = L\,\text{FLOPs}_{\text{LoRA layer}}\]
\label{app:backward flop}

\section{Theoretical Compute Cost of Training meta-Llama-3.2-1B-Instruct} \label{app:cost-meta-llama}

From Appendix ~\ref{app:forward_pass} and Appendix ~\ref{app:backward flop}, and the architectural details presented in Appendix ~\ref{app:meta-llama} and Appendix ~\ref{app:meta-llama-finetune} we can estimate the computational cost of one iteration of forward pass and backpropagation per sample for meta-llama. 
\begin{align*}
\text{FLOPs}^{\text{fwd}}_{\text{block}}
  &= 4m(d^{2} + 2dr) + 2m^{2}d \\
  &\quad
   + m(dd_{\text{ff}} + dr + d_{\text{ff}}r)(n_{f1} + n_{f2}), \\[4pt]
\text{FLOPs}^{\text{bwd}}_{\text{block}}
  &= 4m(d^{2} + 4dr) \\
  &\quad
   + m(n_{f1} + n_{f2})(dd_{\text{ff}} + 2dr + 2d_{\text{ff}}r)\\
   &\quad+ 4m^{2}d, \\[4pt]
\text{FLOPs}_{\text{block}}
  &= \text{FLOPs}^{\text{fwd}}_{\text{block}}
   + \text{FLOPs}^{\text{bwd}}_{\text{block}} \\
  &= 6m^2d + 8md^2 +
  24mdr + \\
   & (n_{f1} + n_{f2})(2mdd_{ff} + 3mdr + 3md_{ff}r) \\[4pt]
 \text{FLOPs}_{\text{model}}
  &= L \cdot \text{FLOPs}_{\text{block}} .
\label{single-pass-flop}
\end{align*}

Plugging-in appropriate values, 
\begin{align*}
\text{FLOPs}_{\text{block}} &= 159,653,888\,m + 12,288\,m^{2} \\
\text{FLOPs}_{\text{model}}&= 2,554,462,208\,m + 196,608\,m^{2} \\
\end{align*}

\section{Theoretical Compute Cost of Training Soft Contextualized Encoder}\label{app:cost-soft-contextualized-encoder}
From Appendix ~\ref{app:forward_pass} and Appendix ~\ref{app:backward flop}, and the architectural details presented in Appendix ~\ref{app:roberta} and Appendix ~\ref{app:jina} we can estimate the computational cost of forward pass and backpropagation per sample for the trainable permutation invariant encoder part, and the cost of forward pass per sample for the embedding generating model part.

Cost for RoBERTa (encoder part) assuming m-1 labels (tokens). Note: the external embedding acts as a pseudo token and there is an additional linear layer that projects the external embedding $d_{\text{ext}}$ to the hidden dimension $d$ ( $\sim$ incurring an additional cost of $3d d_{\text{ext}}$) :
\begin{align*}
\text{FLOPs}^{\text{fwd}}_{\text{block}}
  &= 4m d^{2}
     + 2m^{2} d\\
  & + m d d_{\mathrm{ff}} (n_{f1}+n_{f2})
     , \\[6pt]
\text{FLOPs}^{\text{bwd}}_{\text{block}}
  &= 8m d^{2}
     + 4m^{2} d \\
   & + 2m d d_{\mathrm{ff}} (n_{f1}+n_{f2})
     , \\[6pt]
\text{FLOPs}_{\text{block}}
  &= \text{FLOPs}^{\text{fwd}}_{\text{block}}
   + \text{FLOPs}^{\text{bwd}}_{\text{block}} \\[3pt]
  &= 12m d^{2}
     + 6m^{2} d
     + 3m d d_{\mathrm{ff}}(n_{f1}+n_{f2})\\
\text{FLOPs}_{\text{model}}
  &= L \cdot \text{FLOPs}_{\text{block}}  + 3d d_{\text{ext}} .
\end{align*}
Plugging in appropriate value,
\begin{align*}
\text{FLOPs}_{\text{model}} &= 254,803,968m + 55,296m^2 \\
&+ 2,359,296
\end{align*}

The one off-Inference cost for jina-embedding-v3 (external embedding model part) assuming m tokens/text-sentence excerpt:\footnote{Once calculated, they can be reused for subsequent epochs} \\
\begin{align*}
&\text{FLOPs}_{\text{block}}=\text{FLOPs}^{\text{fwd}}_{\text{block}}\\&
  = 4m d^{2}
     + 2m^{2} d
     + m d d_{\mathrm{ff}} (n_{f1}+n_{f2}), \\[6pt]
&\text{FLOPs}_{\text{model}}
  = L \cdot \text{FLOPs}_{\text{block}}\\[6pt] 
\end{align*}

Plugging in appropriate values:
\begin{align*}
\text{FLOPs}_{\text{model}} = 301,989,888m + 49,152m^2
\end{align*}
As the number of epoch iterations increases, the one off cost of inference becomes less and less significant - with cost increments being only due to the trainable encoder (RoBERTa).
\section{Empirical training time (per epoch) under Experimental Setup}\label{app:Comparison-empirical}
 Training meta-Llama-3.2-1B incurs substantial (NVIDIA RTX A6000-50GB Vram) GPU memory overhead, which restricts the feasible batch size to 2 samples per batch. To reach a reasonable effective batch size, training therefore requires gradient accumulation over 32 steps, resulting in an epoch of 8000 micro batches for N=16000 samples (training dataset size). Though training only involves changing the LoRa adapter parameters (126 million), back propagation and forward pass still involves the whole 1 Billion parameter model.  A single training epoch thus requires approximately 54 minutes or 3240 sec.\\
In contrast, the soft contextualized encoder is significantly more memory efficient, allowing training with 512 samples per batch without gradient accumulation. Training involves exclusively one part of the architecture - tuning the 110 million the RoBERTa (base).As a result, one epoch comprising of 32 batches is completed under 20 seconds on the same GPU (being 0.006x Llama train time).The external embeddings (from jina-embedding v3) can be calculated once, stored and reused. 

\section{Theoretical training time (per epoch) under Experimental Setup} \label{app:Comparison cost}
\paragraph{LoRa-finetuned-meta-Llama-3.2-1B-Instruct.}
We fine-tune meta-llama using the prompt template described in Appendix ~\ref{app:prompt:llama-single}. The template itself, consists of 140 tokens under llama-tokenizer. Each training sample consists of an input text (on average 75 tokens) and a list of candidate categories (on average 10 tokens), resulting in a formatted prompt of approximately 225 tokens. During training the model input takes the form 
\texttt{[formatted prompt]} + \texttt{[correct\_label]} + \texttt{<eot\_id>}, with loss computed only on the label and eot token. This set-up formally involves about two auto-regressive generation steps, however for consistency we conservatively account for a single generation step per sample.
The dataset described in Appendix ~\ref{app:datasets} contains $N=16000$ training samples.
Using per sample compute estimates derived in Appendix ~\ref{app:cost-meta-llama}, the total training cost per epoch is $\text{FLOPs/epoch}= N  \text{FLOPs}_{\text{model}}= 9.355\times 10^{15}$.

\paragraph{Soft Contextualized Encoder.}
For the soft contextualized encoder, the average number of tokens per text sample processed by jina-embedding-v3 is $m_{\text{jina}} =$ 84.17. The average number of tokens (labels)  processed by RoBERTa is \footnote{text-embedding is soft prompted} $m_{\text{Roberta}} - 1 =$ 10. The number of training samples is $N=16000$ (Appendix~\ref{app:datasets}).
Using the compute cost derived in Appendix~\ref{app:cost-soft-contextualized-encoder}, the total training cost per epoch for RoBERTa is  $\text{FLOPs/epoch}$ =$N\text{FLOPs}_{\text{model}}$=$4.5\times 10^{13}$. The one-off cost  incurred for computing jina-embedding-v3 representations is $N\text{FLOPs}_{\text{model}}$=$4.12\times 10^{14}$.

\end{document}